\documentclass[letterpaper]{article} 
\usepackage{aaai2026}  
\usepackage{times}  
\usepackage{helvet}  
\usepackage{courier}  
\usepackage[hyphens]{url}  
\usepackage{graphicx} 
\usepackage{natbib}  
\usepackage{caption} 
\usepackage{algorithm}
\usepackage{algorithmic}
\usepackage{pifont}
\usepackage{newfloat}
\usepackage{listings}
\DeclareCaptionStyle{ruled}{labelfont=normalfont,labelsep=colon,strut=off} 
\floatstyle{ruled}
\newfloat{listing}{tb}{lst}{}
\floatname{listing}{Listing}
\title{SETR: A Two-Stage Semantic‐Enhanced Framework for Zero‐Shot Composed Image Retrieval}
\author{
    Yuqi Xiao,
    Yingying Zhu\equalcontrib,
}
\affiliations{
    \textsuperscript{\rm }College of Computer Science and Software Engineering, Shenzhen University, China\\
zhuyy@szu.edu.cn

}

\usepackage{enumitem}
\usepackage{multirow}
\usepackage[most]{tcolorbox}
\usepackage{subcaption}
\usepackage{booktabs}

\begin{document}

\maketitle

\begin{abstract}
Zero-shot Composed Image Retrieval (ZS-CIR) aims to retrieve a target image given a reference image and a relative text, without relying on costly triplet annotations. Existing CLIP-based methods face two core challenges: (1) union-based feature fusion indiscriminately aggregates all visual cues, carrying over irrelevant background details that dilute the intended modification, and (2) global cosine similarity from CLIP embeddings lacks the ability to resolve fine-grained semantic relations. To address these issues, we propose SETR (Semantic-enhanced Two-Stage Retrieval). In the coarse retrieval stage, SETR introduces an intersection-driven strategy that retains only the overlapping semantics between the reference image and relative text, thereby filtering out distractors inherent to union-based fusion and producing a cleaner, high-precision candidate set. In the fine-grained re-ranking stage, we adapt a pretrained multimodal LLM with LoRA to conduct binary semantic relevance judgments (“Yes/No”), which goes beyond CLIP’s global feature matching by explicitly verifying relational and attribute-level consistency. Together, these two stages form a complementary pipeline: coarse retrieval narrows the candidate pool with high recall, while re-ranking ensures precise alignment with nuanced textual modifications. Experiments on CIRR, Fashion-IQ, and CIRCO show that SETR achieves new state-of-the-art performance, improving Recall@1 on CIRR by up to 15.15 points. Our results establish two-stage reasoning as a general paradigm for robust and portable ZS-CIR.
\end{abstract}
\section{Introduction}
Composed Image Retrieval (CIR) seeks to retrieve a target image \(I_t\) from a large gallery, given a reference image \(I_r\) and a relative text \(T\). This task has significant applications in e-commerce, visual search, and content creation. However, traditional CIR methods require densely annotated triplets \((I_r, T, I_t)\), whose collection is both labor-intensive and domain-specific~\cite{Baldrati2022,Wen2023,Zhang2021}. To circumvent this, Zero-Shot CIR (ZS-CIR) has emerged as a promising paradigm, leveraging powerful pretrained vision-language models (VLMs) like CLIP~\cite{radford2021learning} to eliminate the need for task-specific supervision.

\begin{figure}[ht]
  \centering
  \includegraphics[width=\linewidth]{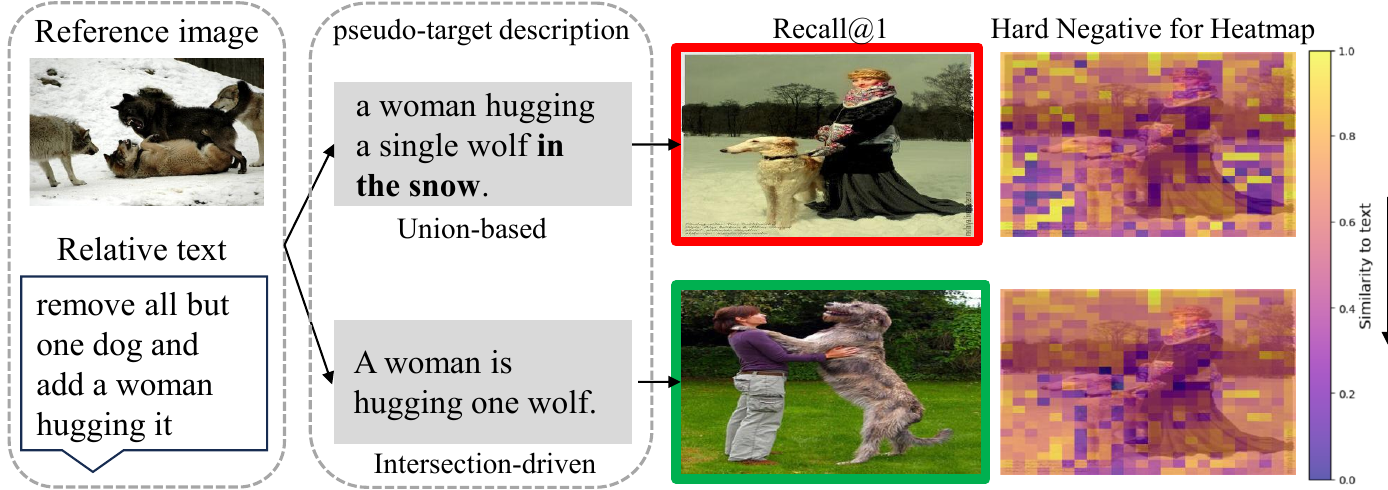}
\caption{Comparison of union-based vs. intersection-driven strategies. We show attention heatmaps on the same hard negative image retrieved by the union-based strategy (red border). Top: the union-based strategy retains the irrelevant “in the snow” cue, ranks the wrong image at recall@1 (red border), and its similarity attention focuses on the snowy background. Bottom: our intersection-driven strategy correctly retrieves the target image (green border) with a clean pseudo-target description, and its similarity attention highlights the precise clue “hugging.”}
  \label{fig:intersection_strategy}
\end{figure}

A dominant design choice in existing ZS-CIR is the union-based fusion strategy, where the visual representation of the reference image is combined with the semantics of the relative text to construct a query~\cite{saito2023pic2word,li2025rethinking,gu2024language,yang2024ldre,tang2024context,tang2024reasonbeforeretrieveonestagereflectivechainofthoughts}. This approach aims to maximize coverage by retaining all cues from the reference image and relative text. However, its inclusiveness is also its weakness: irrelevant background or contextual details from the reference image are indiscriminately merged, leading to semantic interference that often overwhelms the intended modification. Figure~\ref{fig:intersection_strategy} highlights this limitation: Given the query "remove all but one dog and add a woman hugging it,” union-based strategy latches more onto the unrelated semantic snow background rather than the directive "hugging" and fails to retrieve the correct target.

We argue that this union paradigm fundamentally misaligns with the objective of CIR. The task is not to collect all semantics from the reference image but to selectively preserve only what remains unchanged while applying the instructed modification(relative text). Motivated by this insight, we introduce the intersection paradigm for semantic fusion, which explicitly models the overlap \( I_r \cap T \) between the reference image and the relative text. By filtering out the residual set \( I_r \setminus T \), our approach focuses retrieval on target-relevant regions of the reference image and eliminates interference. Unlike prior models that implicitly rely on LLM-generated pseudo-descriptions (e.g., LDRE~\cite{yang2024ldre}), context-dependent pseudo-words (Context-I2W~\cite{tang2024context}), or object-centric tokenization (MOA~\cite{li2025rethinking}), these approaches, despite their surface differences, remain anchored in the union paradigm: their objective is to integrate as much semantic content as possible from the reference and text. This inclusiveness inevitably carries over irrelevant cues, which dilute the intended modification and hinder retrieval precision. By contrast, our intersection-driven strategy pioneers a new paradigm for information fusion—it selectively retains only the overlapping, text-consistent semantics. While this may sacrifice some potentially useful information, it substantially reduces noise and enhances the authenticity of the retained cues. This trade-off yields a more faithful alignment between the composed query and the target image, and, by narrowing the candidate pool, facilitates more effective downstream re-ranking.


Yet, even with cleaner candidates, CLIP’s global cosine similarity struggles to discriminate subtle variations (e.g., distinguishing “a red striped shirt” from “a red plain shirt”). To address this limitation, our SETR (Semantic-enhanced Two-Stage Retrieval) introduces a complementary second stage: an MLLM-based fine-grained scorer. Adapted via Low-Rank Adaptation (LoRA), the MLLM transforms compositional queries and candidate images into binary semantic relevance judgments (“Yes/No”), which we aggregate into structured scores for re-ranking. Unlike CLIP’s coarse global matching, this scorer explicitly verifies fine-grained consistency, yielding more reliable fine-grained alignment. Our design offers both conceptual clarity and portability as a general enhancement to retrieval pipelines.

Together, these two stages form a collaborative pipeline: intersection-driven coarse retrieval narrows the search space with high-precision candidate sets, and MLLM-based re-ranking refines them with nuanced semantic reasoning. This synergy delivers consistent improvements across CIRR, CIRCO, and FashionIQ (Table~\ref{tab:comparison}), establishing intersection-driven fusion as a pioneering paradigm for information alignment in ZS-CIR.

Our main contributions are as follows.
\begin{itemize}[leftmargin=1em]
  \item We introduce a new paradigm for compositional query construction in ZS-CIR. Unlike union-based methods that indiscriminately aggregate all cues, our intersection strategy selectively preserves only overlapping, text-consistent semantics. This reduces interference and enhances authenticity, achieving a precision-oriented trade-off.
  
  \item We adapt a pretrained multimodal LLM with LoRA to perform binary semantic relevance scorer, enabling explicit verification of relational and attribute-level consistency. This provides finer discrimination than CLIP’s global cosine similarity, especially for subtle modifications (e.g., striped vs. plain).
  
  \item We integrate intersection-driven coarse retrieval with MLLM-based re-ranking into the SETR framework. The two stages are complementary: coarse retrieval narrows the candidate pool with high recall, and re-ranking refines candidates with high precision. Experiments on CIRR, CIRCO, and FashionIQ demonstrate consistent state-of-the-art performance, highlighting SETR as a general and portable paradigm for ZS-CIR.

\end{itemize}

\label{sec:introd}
\section{Related Work}

\subsection{Zero-Shot CIR (ZS-CIR)}

Recent advances in ZS-CIR \cite{saito2023pic2word,baldrati2023zero,gu2023compodiff} have eliminated the need for annotated triplets by leveraging pretrained vision–language models such as CLIP. Most approaches follow a \textbf{union-based composition strategy}, combining reference and text features to maximize semantic coverage. However, this inclusiveness inevitably imports irrelevant cues from the reference image, which dilute the intended modification and inject semantic noise into retrieval.

Despite employing different mechanisms, recent methods all share this limitation. SEIZE~\cite{yang2024ldre} use LLM-generated multiple pseudo-descriptions, Context-I2W~\cite{tang2024context} design context-dependent pseudo-words,  MOA~\cite{li2025rethinking} adopt object-centric tokenization, and CoTMR~\cite{sun2025cotmrchainofthoughtmultiscalereasoning} use CoT to reason about all possible information, each of these strategies ultimately aggregates information indiscriminately under the union paradigm, leaving semantic interference unresolved. 

In contrast, our intersection-driven strategy introduces a new paradigm for information fusion: rather than integrating all cues, it selectively preserves only the overlapping, text-consistent semantics. This reduces interference and enhances the authenticity of the retained cues, establishing a cleaner foundation for fine-grained re-ranking.

\begin{figure*}[h]
    \centering
    \includegraphics[width=0.9\textwidth]{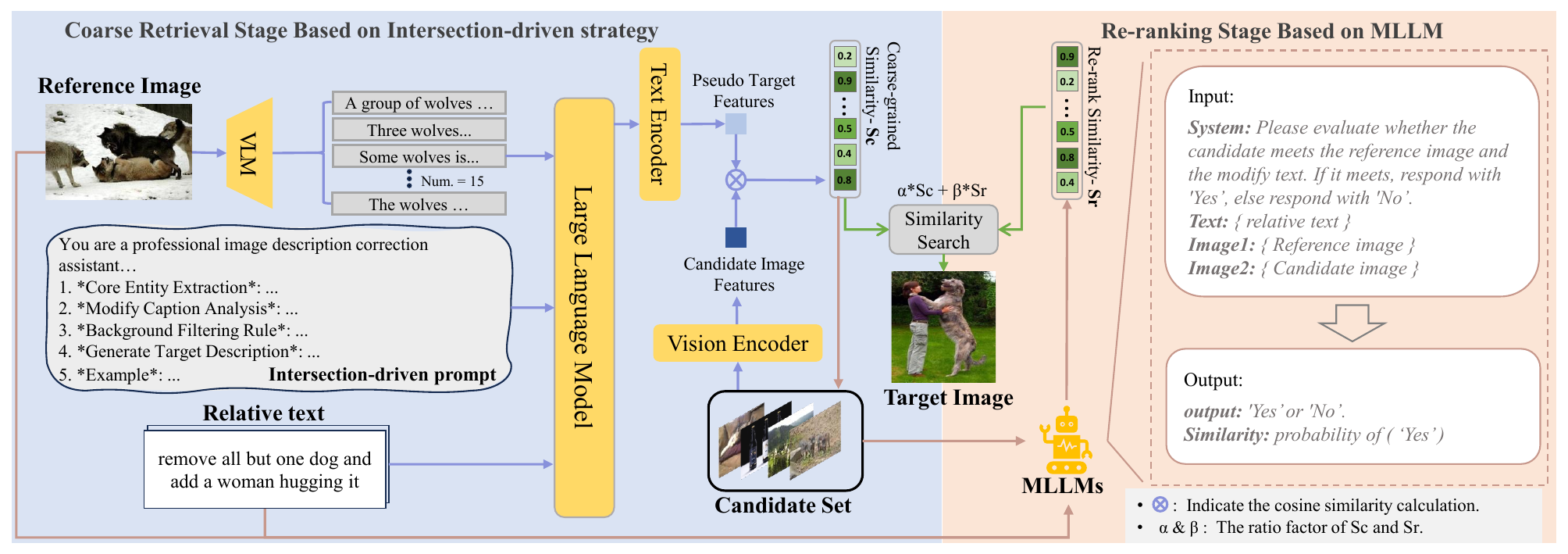}
    \caption{Overall framework of the proposed SETR for ZS-CIR.}

    \label{fig:framework}
\end{figure*}
\subsection{Vision–Language Models and MLLMs}

Vision–Language Models (VLMs) such as UNITER \cite{chen2020uniter} and CLIP \cite{radford2021learning} have become central to multimodal representation learning by aligning image–text pairs in a shared embedding space. CLIP, in particular, achieves strong performance in zero-shot tasks and is widely adopted for CIR.

Beyond VLMs, the rise of Multimodal Large Language Models (MLLMs) such as GPT‑4o \cite{openai2024gpt4ocard} and Qwen2.5‑VL \cite{bai2025qwen25vltechnicalreport} enables fine-grained cross-modal reasoning. Unlike CLIP’s global feature matching, MLLMs incorporate visual inputs into a language-centric reasoning, allowing them to interpret nuanced relational semantics \cite{liu2024lamralargemultimodalmodel}. Our method leverages this capability in a two-stage pipeline: coarse filtering stage using CLIP, followed by semantic re-ranking powered by MLLMs.

\subsection{Cosine Similarity}

Cosine similarity is a popular metric in CIR due to its simplicity and compatibility with contrastively trained models like CLIP. However, it assumes isotropic embeddings and cannot adequately model fine-grained semantic changes. This limitation is especially pronounced in ZS-CIR, where visually similar yet semantically incorrect images are often retrieved \cite{yang2024ldre}.
 
Recent work \cite{Sun2025CoTMR,Borgeaud2023Retro} shows that replacing static similarity metrics with reasoning-capable models can alleviate this issue. MLLMs offer context-aware interpretation of image–text pairs and support dynamic understanding of compositional instructions. Motivated by this, our method integrates classical similarity for coarse candidate generation with MLLM-based semantic reasoning for refined ranking, enabling more accurate retrieval under complex transformations.

\section{Method}

\subsection{Overview}

We propose SETR, a two-stage framework. As illustrated in Figure~\ref{fig:framework}, SETR consists of: (1) coarse retrieval stage employing an intersection-driven strategy to prune the candidate set, and (2) fine-grained re-ranking stage leveraging a fine-tuned MLLM scorer for semantic consistency scoring.

\subsection{Intersection-driven Coarse Retrieval stage}
\paragraph{Intersection-driven strategy.}
To improve the semantic alignment of the pseudo-target description, we introduce a intersection-driven strategy that identifies the core semantic intersection between the reference image \( I_r \) and the relative text \( T \), denoted as \( I_r \cap T \).\footnote{\( I_r \cap T \) represents the shared visual semantic between the reference image and the relative text.} This retains only the visual elements from \( I_r \) that are intended to persist in the target image, minimizing semantic drift and irrelevant carry-over.

We additionally identify and discard visual elements in \( I_r \) that are not clear supported by \( T \), represented by the set difference \( I_r \setminus T \).\footnote{\( I_r \setminus T \) refers to visual content in the \( I_r \) that contradicts or is irrelevant to the relative text.} Removing these mitigates interference information and ambiguity from overdescriptive captions, thereby sharpening the retrieval focus.

To implement this semantic filtering, we design a prompt-based framework that guides to reason jointly over visual description from the reference image and modification request from the relative text. Inspired by prior work on prompt design \cite{zhao2021calibrateuseimprovingfewshot}, we propose a modular engineering to ensure semantic extraction across diverse CIR scenarios. The resulting pseudo-target descriptions are used as text queries during coarse retrieval, improving matching precision.

The prompt content of the Perception enhancement step is as follows. Note that we provide the reference image and relative text in the context and refer to it accordingly in the prompt description.
\begin{tcolorbox}[breakable, colframe=black, colback=white, sharp corners=south, width=\columnwidth, boxrule=0.5mm]
\textbf{Perception Enhancement Prompt Framework} \\
\small
\texttt{You are a professional image description correction assistant. Please process the input information according to the following rules:}
\begin{enumerate}[leftmargin=1.5em]
    \item \texttt{Core Entity Extraction: Identify main entity types (e.g., person, object) on the Image Description from user input:[Image Content]......}
    \item \texttt{Relative Text Analysis: The modification Caption from user input is:[Instruction]......}
    \item \texttt{Background Filtering: Delete the content that......}
    \item \texttt{Target Description Generation: The description will be kept as simple and concise as possible. It will focus solely on the primary subject or action outlined in the Instruction.......}
    \item \texttt{Example: Image Description:"The image shows a woman standing on top of a lush green field,holding a large Irish Wolfhound in her arms." Caption:"shows two dogs eating." Target Description:"Two dogs eating."}
\end{enumerate}
\end{tcolorbox}

\paragraph{Coarse Retrieval with CLIP.}
Given a gallery \( B = \{ I_i \}_{i=1}^M \) and a query in the form of a pseudo-target text, we compute normalized CLIP embeddings:
\begin{equation}
    \mathbf{v}_i = \frac{\psi_I(I_i)}{\|\psi_I(I_i)\|_2}, \quad 
    \mathbf{t}   = \frac{\psi_T(T)}{\|\psi_T(T)\|_2},
\end{equation}
where \( \psi_I \) and \( \psi_T \) are the CLIP image and text encoders, respectively. The cosine similarity is computed as Equation~\eqref{eq:cs}:
\begin{equation}
    S_c(I_i, T) = \mathbf{v}_i^\top \mathbf{t}.
\label{eq:cs}
\end{equation}
Sorting \( \{ S_c(I_i, T) \} \) in descending order yields a permutation \( \sigma \), we select the top \( K = 50 \) images as Equation~\eqref{eq:coarse}:
\begin{equation}
    C_{\text{coarse}} = \{ I_{\sigma(1)}, \dots, I_{\sigma(K)} \}.
\label{eq:coarse}
\end{equation}
This step efficiently reduces the retrieval space while retaining semantically plausible candidates for re-ranking.

\subsection{ MLLM-based Fine-grained Re-ranking stage}
To refine the coarse candidate set, we introduce an MLLM-based scorer that converts multimodal reasoning into structured retrieval scores. Instead of relying on CLIP’s global cosine similarity, the scorer evaluates each candidate through binary semantic relevance judgments (“Yes/No”) with respect to the composed query. This scorer explicitly verifies fine-grained semantic consistency, providing a principled scoring mechanism for fine-grained re-ranking.

To adapt the scorer to this binary judgment task, we apply lightweight LoRA adapters, fine-tuning only 1–2\% of parameters while keeping the backbone frozen. Training pairs are constructed from the Visual Genome corpus\cite{krishna2017visual} by combining annotated relation triplets (subject, predicate, object) into complex region-level captions. Importantly, we do not construct retrieval triplets, and all images used in fine-tuning are strictly disjoint from our test sets. The purpose of this adjustment is not to learn retrieval supervision, but to stabilize the scoring mechanism, ensure reliable yes/no consistency judgments, and improve the output speed of scoring results through task adaptation. Thus, the retrieval stage remains strictly zero-shot, with re-ranking implemented as a portable scorer module.

At inference time, the scorer assigns a semantic-consistency score to each candidate independently. For efficiency, we default to re-ranking the top \( K = 10 \) candidates from the coarse retrieval stage.

\begin{table*}[h]
    \centering
    \resizebox{0.95\textwidth}{!}{%
    \begin{tabular}{llccccccccccc}
        \toprule
        \multirow{3}{*}{Backbone} & \multicolumn{1}{c}{\multirow{3}{*}{Method}} & \multicolumn{1}{c}{\multirow{3}{*}{LLM-based}} &  \multicolumn{4}{c}{CIRCO} & \multicolumn{5}{c}{CIRR} & \multicolumn{1}{c}{FashionIQ}  \\
        \cmidrule(l){4-7} \cmidrule(l){8-12}\cmidrule(l){13-13}
        & & &  \multicolumn{4}{c}{mAP@k}  &  \multicolumn{3}{c}{Recall@k} & \multicolumn{2}{c}{Rs@k} & \multicolumn{1}{c}{R@k(avg)}\\
        & & & k=5 & k=10 & k=25 & k=50 & k=1 & k=5 & k=10 & k=1  & k=2 & k=10\\
        \midrule
        \multirow{8}{*}{ViT-L/14} 
        & Pic2Word(CVPR’23) \dag & \ding{55} &  8.55 & 9.51 & 10.64 & 11.29 & 23.90 & 51.70 & 65.30 & 53.76 & 74.46 & 24.70  \\
        & LinCIR(CVPR’24) \dag & \ding{55} &  12.59 & 13.58 & 15.00 & 15.85 & 25.04 & 53.25 & 66.68 & 57.11 & 77.37 & 26.28 \\
         & Context-I2W(AAAI’24) \dag & \ding{55} &  13.04 &  14.62 & 16.14 & 17.16 & 25.60 & 55.10 & 68.50 & - & - & 27.80 \\
        & MOA(SIGIR’25) \dag & \ding{55} &  15.30 &  17.10 &  18.50 & 19.30 & 27.10 & 56.50 & 69.20 & - & - & 30.10 \\
        & CIReVL(ICLR’24) \dag & \ding{51} &  18.57  & 19.01  & 20.89  & 21.80  & 24.55  & 52.31  & 64.92  & 59.54  & 79.88 & 28.55 \\
        & LDRE(SIGIR’24) \dag & \ding{51} &  23.35  & 24.03  & 26.44  & 27.50  & 26.53  & 55.57  & 67.54  & 60.43  & 80.31 & 24.81 \\
        & OSrCIR(CVPR’25) \dag & \ding{51} &  \underline{23.87}  & \underline{25.33}  & \underline{27.84} & \underline{28.97} & \underline{29.45} & \underline{57.68} & \underline{69.86} & \underline{62.12} & \underline{81.92} & \underline{33.26} \\
        & SETR (Ours) & \ding{51} &  \textbf{30.87} & \textbf{29.65} & \textbf{31.70} & \textbf{32.73} & \textbf{41.68} & \textbf{66.53} & \textbf{71.76} & \textbf{72.17} & \textbf{87.52} & \textbf{35.32} \\
        \midrule
        \multirow{4}{*}{ViT-G/14} 
        & Pic2Word(CVPR’23) \dag & \ding{55} &  5.54 & 5.59 & 6.68 & 7.12 & 30.41 & 58.12 & 69.23 & 68.92 & 85.45 & 31.28  \\
        & LinCIR(CVPR’24) \dag & \ding{55} &  19.71 & 21.01 & 23.12 & 24.18 & 34.80 & 64.07 & 75.11 & 68.72 & 84.70 & \textbf{45.11} \\
        & CIReVL(ICLR’24) \dag & \ding{51} &  26.77  & 27.59  & 29.96  & 31.03  & 34.65  & 64.29  & 75.06  & 67.95  & 84.87 & 32.19 \\
        & LDRE(SIGIR’24) \dag & \ding{51} &  \underline{31.12} & \underline{32.24} & 34.95 & 36.03 & 36.15 & 66.39 & 77.25 & 68.82 & \underline{85.66} & 32.49 \\
        & OSrCIR(CVPR’25) \dag & \ding{51} &  30.47 & 31.14 & \underline{35.03} & \underline{36.59} & \underline{37.26} & \underline{67.25} & \underline{77.33} & \underline{69.22} & 85.28 & 37.57\\
        & SETR (Ours) & \ding{51} &  \textbf{37.98} & \textbf{37.72} & \textbf{40.34} & \textbf{41.37} & \textbf{44.36} & \textbf{73.13} & \textbf{79.71} & \textbf{75.66} & \textbf{89.45} & \underline{37.87} \\
        \bottomrule
    \end{tabular}}
    \caption{Results on ZS-CIR benchmarks CIRCO, CIRR and FashionIQ. The top two scores are emphasized, with the highest in bold and the runner-up underlined. (\dag indicates the results are cited from \cite{tang2024reasonbeforeretrieveonestagereflectivechainofthoughts,li2025rethinking,yang2024semantic}.)}
    \label{tab:comparison}
\end{table*}

\paragraph{Semantic Representation Construction.}  
For each candidate image \( c_i \in C_{\text{top10}} \), we construct an input \( P_i \) by encoding the visual features from \( c_i \) and the combined context of the relative text and reference image, as shown in Equation~\eqref{eq:pi}:
\begin{equation}
P_i = \Phi_{\text{vis}}(c_i) \oplus \Phi_{\text{text}}(T, I_{\text{r}}),
\label{eq:pi}
\end{equation}
where \( \Phi_{\text{vis}} \) and \( \Phi_{\text{text}} \) are the image and text encoders of the MLLM, and \( \oplus \) denotes multimodal fusion.

\paragraph{Consistency Scoring.}  
The scorer is prompted to assess the relevance of each \( P_i \) and outputs probability for 'yes' options, as shown in Equation~\eqref{eq:sc}:
\begin{equation}
S_r(c_i) = \text{Probability}(\text{MLLM}_{\text{YES}}).
\label{eq:sc}
\end{equation}
This score reflects the model’s confidence that candidate \( c_i \) semantically matches the intended transformation.

\paragraph{Dynamic Re-ranking.}  
We combine the original CLIP cosine similarity \( S_c \) and the MLLM consistency score \( S_r \) to produce a final relevance score as Equation~\eqref{eq:final}:
\begin{equation}
\begin{aligned}
L_{\text{final}} 
= \text{SORT}\Big(
    &\{ \alpha \cdot S_c(j) + \beta \cdot S_r(j) \}_{j=1}^{10} \\
    &\cup \{ S_c(j) \}_{j=11}^{50} 
\Big),
\end{aligned}
\label{eq:final}
\end{equation}
where \( \alpha \) and \( \beta \) control the weighting between coarse similarity and re-ranking score, and \( \cup \) denotes set concatenation. This late fusion strategy balances efficiency and accuracy by applying MLLM scoring only to the top-ranked candidates.

\section{Experiment}
\subsection{Experimental Setup}
\textbf{Datasets:} Researchers widely use CIRR\cite{liu2021image}, FashionIQ\cite{guo2019fashion}, and CIRCO\cite{Baldrati2022} for compositional image retrieval, each targeting different aspects of the task. CIRR emphasizes complex real-world semantics, including attribute variations, spatial relations, and compositionality. CIRCO introduces a one-to-many retrieval challenge that requires the model to reason about semantics while filtering out irrelevant noise. FashionIQ focuses on modifying specific attributes of images in the fashion field. CIRR and CIRCO datasets focus on semantically rich natural world images, which are more complex and realistic, and also include the retrieval of fashion information to a certain extent. Therefore, we focus on evaluating the performance changes on CIRCO and CIRR datasets.

\textbf{Evaluation Metrics:} We evaluate retrieval performance using Recall@K (R@K) for one-to-one settings on FashionIQ and CIRR. For CIRR, we additionally report $Recall_{\text{subset}}@K$ (Rs@K) on a subset where each query is matched against six semantically similar candidates. For the one-to-many scenario in CIRCO, we use mean average precision (mAP), which captures both relevance and ranking quality.

\textbf{Implementation Details:} Following the findings of Yang et al., who have analyzed various configurations of VLMs and LLMs, we set the number of reference captions to 15 for optimal performance\cite{yang2024ldre}. We use CLIP with ViT-L/14 and ViT-G/14 backbones as the vision–language model, and GPT-4o~\cite{openai2024gpt4ocard} as the language model, consistent with baseline settings to ensure fair comparison, and Qwen2.5-VL~\cite{bai2025qwen25vltechnicalreport} as the MLLM for task-oriente fine-tuning. During re-ranking, we train for one epoch on a single A100 GPU and re-ranking for the top-10 coarse retrieval candidates due to computational cost.

\subsection{Comparison with SOTA Methods}

We compared SETR with state-of-the-art (SOTA) methods on ZS-CIR benchmark. These methods include \textbf{Pic2Word}\cite{saito2023pic2word}, \textbf{LinCIR}\cite{gu2024language}: maps the visual features of a reference image into a pseudo-word token, \textbf{Context-I2W}\cite{tang2024context}: selectively
maps text description-relevant visual information from the
reference image,  \textbf{MOA}\cite{li2025rethinking}: through object recognition and text-based object screening, the objects to be modified in the reference image are adaptively converted into pseudo-words, \textbf{OSrCIR}\cite{tang2024reasonbeforeretrieveonestagereflectivechainofthoughts}: use the reflective thought chaining framework to infer pseudo-target descriptions by combining relative text with contextual clues in the reference image, as well as  \textbf{CIReVL}\cite{karthik2024visionbylanguagetrainingfreecompositionalimage} and the baseline model \textbf{LDRE}\cite{yang2024ldre}: the LLM-based method on the union-based strategy obtains the pseudo-target image description.

Table~\ref{tab:comparison} reports evaluation results on ZS‑CIR benchmarks. Our method consistently outperforms all recall metrics, achieving state‑of‑the‑art results. On CIRR with a ViT‑L/14 backbone, we improve Recall@1 by 15.15 percentage points(pp) over LDRE and 12.23 pp over the prior SOTA (OSrCIR), even surpassing the gains of ViT‑G/14 on previous SOTA model. On CIRCO, we boost mAP@5 by 7.52 pp versus the baseline and 7.00 pp over the previous SOTA. These one‑to‑one and one‑to‑many improvements demonstrate the efficacy of our approach in ZS‑CIR.

Our method slightly surpasses prior SOTA on FashionIQ thanks to fine‑grained semantic alignment, though gains are limited by the dataset’s simple attribute edits. With ViT‑L/14, we match or exceed top methods; with ViT‑G/14, we place second behind LinCIR, which benefits from explicit CLIP alignment training. This gap likely reflects LinCIR’s specialized alignment procedure, which our framework lacks—an important limitation in the fashion domain where base CLIP’s pretraining may not capture terms like “sequin corset.”  In contrast, on natural‑image benchmarks like CIRCO, our method outperform all text‑inversion techniques. Future work will explore tighter coupling between reasoning and retrieval to further improve domain‑specialized performance.    

\begin{table}[h]
\centering
\resizebox{0.95\linewidth}{!}{%
\begin{tabular}{lc}
\toprule
Method Variant                                   & R@5 \\
\midrule
1. \textit{Reported Baseline from LDRE \cite{yang2024ldre}} & 66.39 \\
\midrule
2. Reproduced Baseline       & 62.46 \\
\textbf{Significance of key modules of SETR} \\
3. Baseline + Intersection-driven strategy   & 68.87 \\
4. Baseline + MLLM-based Re-ranking  & 68.31 \\
\textbf{Impact of different MLLM backbone size} \\
5. w/ Qwen2.5-VL 3B backbone & 69.88 \\
6. w/ Qwen2.5-VL 7B backbone &  \textbf{73.13} \\
\midrule
7. SETR (ours)             & \textbf{73.13} \\
\bottomrule
\end{tabular}}
\caption{Ablation results on the CIRR test set. See section~\ref{sec:ablation} for more details of method variants settings.}
\label{tab:ablation}
\vspace{-1em}
\end{table}

\subsection{Ablation Studies}
\label{sec:ablation}
To assess the contribution of each component in SETR, we conduct a series of ablation experiments on the CIRR test set, and report Recall@5 results in Table~\ref{tab:ablation}. All models are based on CLIP with a ViT-G/14 backbone. Minor differences between our reproduced baselines and previously reported values may stem from variations in hyperparameter tuning or training settings.   \\
\textbf{Significance of key modules.} Model "3.", which omits the MLLM-based re-ranking module, suffers a 4.26 pp performance drop relative to the full model. Model "4.", which replaces our intersection-driven strategy with a union-based baseline strategy, incurs a 4.82 pp decrease—underscoring the essential contributions of both components. \\
\textbf{Impact of MLLM Backbone Size.} Replacing the Qwen2.5-VL 7B model with its smaller 3B variant (model "5.") causes a 3.25 pp decrease. Nevertheless, the 3B variant still outperforms the baseline union-based fusion model (62.46 R@5) by 10.67 pp, demonstrating that even compact MLLMs contribute meaningfully to semantic refinement, with larger models offering further gains.\\
This trend confirms that our method effectively suppresses irrelevant background while preserving target-relevant semantics, especially in challenging scenarios. These findings also provide more concrete and quantitative explanation for semantic benefits of intersection-driven filtering strategy. 

\begin{table}[h]
    \centering
    \resizebox{0.95\linewidth}{!}{%
    \begin{tabular}{cccccc}
    \toprule
        \multicolumn{1}{c}{\multirow{2}{*}{Prompt Mode}} & \multicolumn{3}{c}{Recall@K} & \multicolumn{2}{c}{Rs@K}   \\
        \cmidrule(r){2-4} \cmidrule(l){5-6}
        &  k=1  & k=5 & k=10 & k=1  & k=2   \\
        \midrule
         Union-based strategy & 33.66 & \underline{62.46} & 73.95 & 63.37 & 82.02 \\
         LLM‑Generated  & \underline{33.88} & 61.66 & \underline{74.00} & \underline{64.72} & \underline{82.10}  \\
         Instructed‑Filtered  & 32.24 & 61.49 & 73.54 & 63.61 & 82.07   \\
         Ours & \textbf{39.23}  & \textbf{68.87} & \textbf{79.71}  & \textbf{72.80}  & \textbf{88.08}  \\
    \toprule
    \end{tabular}}
    \caption{Comparison of prompt modes. “LLM‑Generated” denotes the template produced by the LLM when conditioned only on the reference image and its relative text to generate a pseudo‑target description. “Instructed‑Filtered” additionally guides the LLM to filter out irrelevant background details from the reference image.}
    \label{tab:ab2}
\end{table}

\begin{figure*}[ht]
\center
\includegraphics[width=0.9\textwidth]{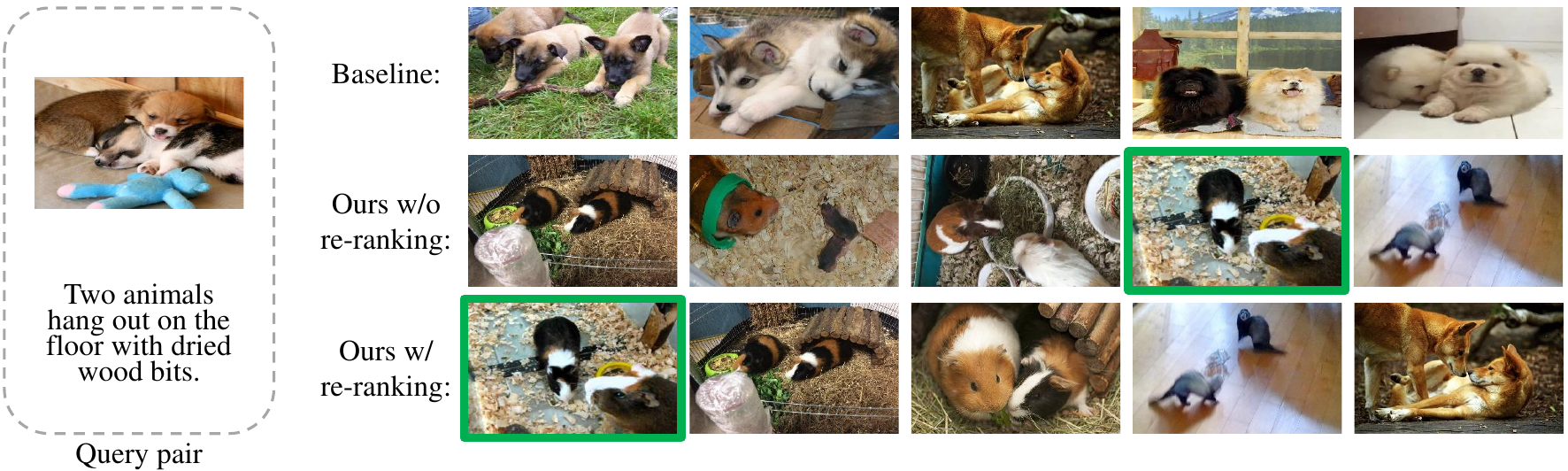}
\caption{Examples from CIRR showing SETR’s improvement over union-based baselines. Our intersection-driven retrieval eliminates background noise, and re-ranking resolves compositional phrasing to recover the correct target.}
\label{fig:quality}
\end{figure*}

\subsection{Prompt Performance Comparison}
To evaluate the Prompt Framework driven by intersection, we compare different prompt configurations in CIRR using Recall@5 (Table~\ref{tab:ab2}). The unstructured baseline prompt ("Union-based strategy") yields 62.46. the LLM-Generated Prompt scores 61.66 (“LLM‑Generated Prompt”), and adding instructions to LLM to filter out irrelevant background details (“Instructed‑Filtered Prompt”) results in 61.49. By contrast, our prompts achieve 68.87, a 6.41 pp gain over baseline and 7.21–7.38 pp over LLM‑Generated variants. These results confirm that our intersection-driven strategy is a important improvment in filter out irrelevant background details, and systematic prompt engineering—not merely incidental wording—substantially enhances the fidelity of pseudo‑target descriptions and drives downstream retrieval performance in ZS-CIR.



\subsection{Qualitative Analysis}
Figure~\ref{fig:quality} illustrates SETR’s two-stage collaboration on a CIRR query where the reference image contains multiple animals and the text specifies “two animals.” The union‑based baseline mistake retrieves dog‑centric results due to interference in reference image and fails to isolate the intended pair. In contrast, SETR’s intersection‑driven coarse retrieval filters out inteference information, bringing the ground truth into the top‑5 candidates, and its MLLM‑based re‑ranking then interprets the relation “on the floor with dried wood bits,” correctly promoting the ground truth to top‑1. This example demonstrates the effectiveness of semantic filtering, the MLLM’s fine‑grained relational reasoning, and the synergy between SETR’s coarse retrieval and re-ranking stages.

\section{Conclusion and Limitations}
Despite clear gains of SETR, it still faces practical hurdles. First, the MLLM-based re-ranking module adds nontrivial latency, hindering real-time or edge deployment. Second, our hard intersection-driven filter can over-prune: subtle, occluded, or context-dependent attributes may fall outside prompt, causing occasional recall misses. Third, SETR’s performance depends on prompt design—small wording or template shifts can induce subtle retrieval variance—and its zero-shot transfer to sharply different domains (e.g., medical imagery, remote sensing) remains unverified without prompt or lightweight fine-tuning.

We have presented SETR, a two-stage ZS-CIR framework that first reduces the search space via intersection-driven filtering and then refines results through MLLM-based re-ranking, achieving state-of-the-art performance on CIRR, FashionIQ, and CIRCO. Going forward, we will pursue lightweight reranking alternatives, develop context-aware filtering to recover nuanced semantics—all while striving to reduce inference cost and improve practical applicability.

\bibliography{aaai2026}

\end{document}